\documentclass[letterpaper, 10 pt, conference]{ieeeconf}  

\IEEEoverridecommandlockouts                              
\usepackage{graphics} 
\usepackage{epsfig} 
\usepackage{amsmath} 
\usepackage{amssymb}  

\usepackage{url}
\usepackage[ruled, vlined, linesnumbered]{algorithm2e}
\usepackage{verbatim} 
\usepackage{soul, color}
\usepackage{lmodern}
\usepackage{fancyhdr}
\usepackage[utf8]{inputenc}
\usepackage{fourier} 
\usepackage{array}
\usepackage{makecell}

\usepackage[utf8]{inputenc} 
\usepackage[T1]{fontenc}    
\usepackage{hyperref}       
\usepackage{url}            
\usepackage{booktabs}       
\usepackage{amsfonts}       
\usepackage{nicefrac}       
\usepackage{microtype}      
\usepackage{lipsum}
\usepackage{graphicx}
\usepackage{float}
\usepackage{balance}

\usepackage{hyperref}
\hypersetup{
    colorlinks=true,
    linkcolor=blue,
    filecolor=magenta,      
    urlcolor=black,
}

\SetNlSty{large}{}{:}

\makeatletter

\newcommand{\Rom}[1]{\expandafter\@slowromancap\romannumeral #1@}
\makeatother

\title{\LARGE \bf
ABD-Net: Attention Based Decomposition Network for 3D Point Cloud Decomposition
}

\author{Siddharth Katageri, Shashidhar V Kudari, Akshaykumar Gunari, Ramesh Ashok Tabib, Uma Mudenagudi
\\ Center of Excellence in Visual Intelligence \\
KLE Technological University \\
Hubli, Karnataka, India \\
{\tt\small siddharth.katageri19@gmail.com, shashidharvk100@gmail.com, akshaygunari@gmail.com, }\\
{\tt\small ramesh\_t@kletech.ac.in, uma@kletech.ac.in}

}

\begin{document}

\maketitle
\thispagestyle{plain}
\pagestyle{plain}

\begin{abstract}
In this paper, we propose Attention Based Decomposition Network (ABD-Net), for point cloud decomposition into basic geometric shapes namely, plane, sphere, cone and cylinder. We show improved performance of 3D object classification using attention features based on primitive shapes in point clouds. Point clouds, being the simple and compact representation of 3D objects have gained increasing popularity. They demand robust methods for feature extraction due to unorderness in point sets. In ABD-Net the proposed Local Proximity Encapsulator captures the local geometric variations along with spatial encoding around each point from the input point sets. The encapsulated local features are further passed to proposed Attention Feature Encoder to learn basic shapes in point cloud. Attention Feature Encoder models geometric relationship between the neighborhoods of all the points resulting in capturing global point cloud information. We demonstrate the results of our proposed ABD-Net on ANSI mechanical component and ModelNet40 datasets. We also demonstrate the effectiveness of ABD-Net over the acquired attention features by improving the performance of 3D object classification on ModelNet40 benchmark dataset and compare them with state-of-the-art techniques.
\end{abstract}


\section{Introduction}
In recent days 3D point cloud is making its ground in every field like, CAD modeling, 3D printing, AR/VR entertainment and self driving cars. There is a need for methods to analyze, process and derive this huge volume of 3D point clouds efficiently. A basic capability of human visual system is to derive relevant structures and their relation from 3D objects. Unlike human vision, supervising a machine to derive such geometrical information is a challenging task. However, representing a 3D object with a set of basic geometric parts simplifies its geometric surface. This simpler representation of a 3D object is vital for better shape understanding, shape information processing and shape analysis tasks.

\begin{figure}
        \includegraphics[ width = 1.0\linewidth]{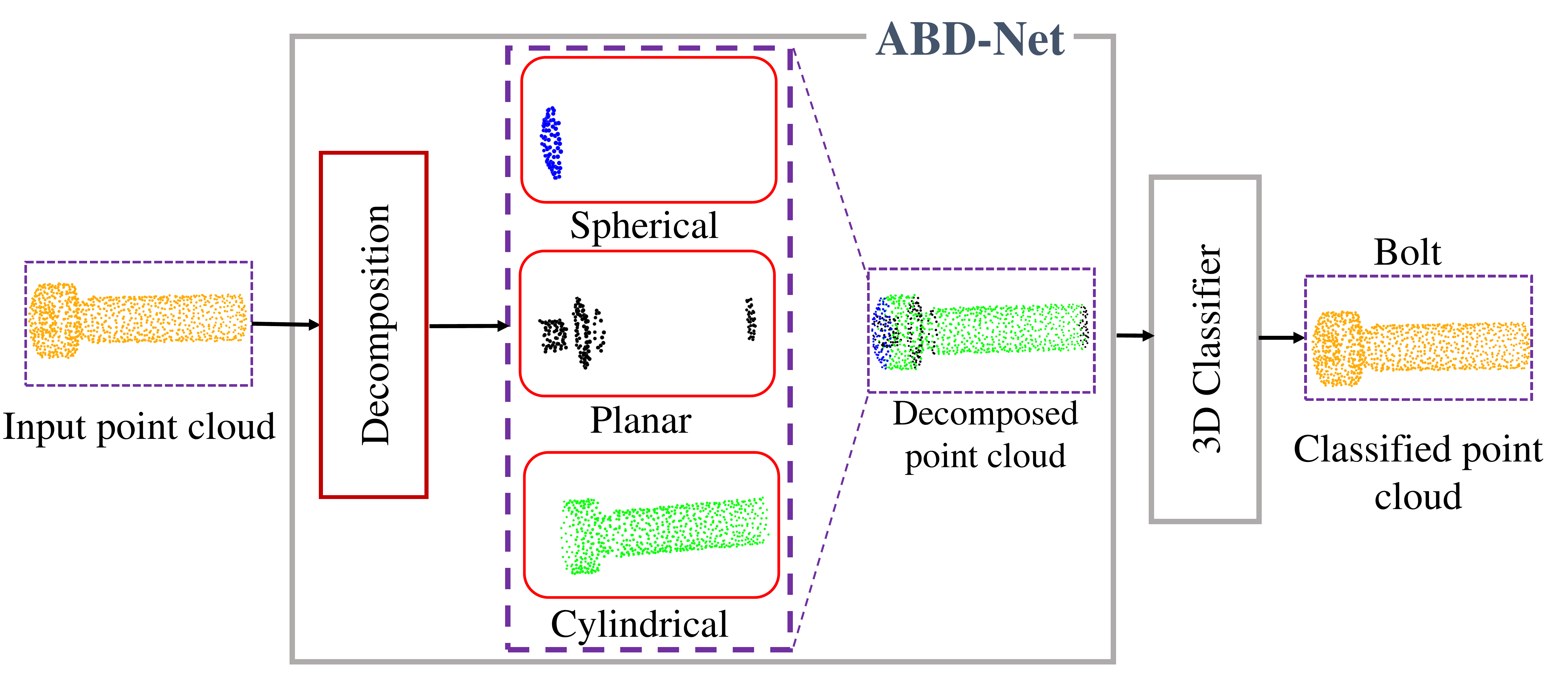}
        \caption{Overview of Attention Based Decomposer (ABD-Net) where basic shape features are used for 3D object classification. Point color indicates the shape to which it belong. [Black for Planar, Blue for Spherical and Green for Cylindrical].}
        \label{fig:teaser_diagram}
\end{figure}

Unlike images, which have a defined regular grid, 3D point clouds are irregular and unordered, restricting the direct use of standard and powerful convolution techniques. Some point set processing approaches transform 3D points to voxel grid representations \cite{point_trans_nov_6}, \cite{point_trans_nov_7} or image projections \cite{point_trans_nov_8}, \cite{point_trans_nov_9}. However, this transformation leads to loss of information and also suffers from high processing complexity. To address these issues, considerable amount of work has been done on point based methods that directly act on 3D points. The main idea is to process each point individually using many filters of unit size, and sharing these filters amongst all points capturing point set features \cite{pointnet}, \cite{pointnet++}. However, these approaches use down-sampling step for defining local neighborhood, which causes costly point correspondence search during interpolation. However, if the task demands to estimate point features for all the original number of points, down-sampling might 
hinder the feature representation of the 3D object.

To address the issue of lack of surfacial information in 3D point clouds, some works try to decompose a 3D point cloud into meaningful parts and try to infer a topological graph by modeling relations between these parts \cite{preceptual_shape_decomposition}. However, this decomposition is highly dependent on the perceptual points selected. While some approaches try to reconstruct a parametric form of the input point clouds, considering fixed geometric shapes \cite{spfn}, \cite{parsenet}. This task of fitting basic geometric shapes suffers from high computational complexity.

In this work, we tackle the problem of missing surfacial information by representing the inherent geometry of 3D point cloud using $4$ basic shapes namely, plane, sphere, cone and cylinder. We perform decomposition by assigning per-point labels from one of the four shapes, making it a simpler problem than basic shape fitting. Towards this, an essential factor is, learning the local topological information of the point sets as we want to represent a 3D object at its lowest geometric abstraction. To address this, we propose Local Proximity Encapsulator, a permutation invariant module, which encapsulates both local geometric variation and spatial encoding around each point. Here, to overcome the limitation of costly point correspondence search, we use k-nearest neighbors around each point to define the local neighborhood. However, learning only these local topological information is not sufficient. The relationship between local neighborhood should not be overlooked, as they are not independent, but represent a whole 3D object. This problem is similar to that of in Natural Language Processing (NLP) where in, relationship between words is to be modeled \cite{attention_is_all_you_need}, \cite{point_trans_dec_41}, \cite{point_trans_dec_4}, \cite{point_trans_dec_5}.
We propose Attention Feature Encoder, to model relationships between neighborhood and capture the underlying shape of the whole 3D object. We use these two modules to train a 3D object decomposer which is vital for any shape understanding and shape analysis tasks. We show improved 3D object classification performance by using ABD-Net as a pre-processing step and provide extensive study.

To summarize, the main contributions of our work are as follows:
\begin{itemize}
    \item We propose ABD-Net that captures the inherent geometry of a 3D point cloud and represents it using basic shapes namely, plane, sphere, cone and cylinder helping various 3D analysis tasks. Towards this:
    \begin{itemize}
        \item We propose Local Proximity Encapsulator (LPE) to capture local geometry with spatial encoding around each point, thus incorporating \emph{local attention}.
        \item {We propose Attention Feature Encoder (AFE) to learn basic shapes in point cloud by modeling geometric relationship between the neighborhood of all the points, and call this as \emph{global attention}} which is based on basic shapes.
    \end{itemize}
    \item We train ABD-Net to learn basic shape features using ANSI mechanical components dataset, which has shape labels assigned to each point in a point cloud. We use these features on a different dataset for a different task, specifically 3D classification.
    \begin{itemize}
        \item We evaluate the performance of proposed ABD-Net for decomposition task on ANSI mechanical components dataset achieving an accuracy of 99.3\%.  
        \item We test the proposed ABD-Net on ModelNet40 dataset and demonstrate the performance of point cloud decomposition.
    \end{itemize}
    \item We show effectiveness of the ABD-Net by showing improved classification performance of a 3D classifier having $3$ times less trainable parameters than state-of-the-art and achieve comparable results.
    \item We provide exhaustive evaluation and ablation study to demonstrate the effectiveness of ABD-Net for both, decomposition and classification.
\end{itemize}

The organization of this paper is as follows. In Section \ref{section:related_works}, we study various methods proposed for point cloud processing and point cloud decomposition as a part of literature review. In Section \ref{abd-net}, we discuss the proposed ABD-Net architecture that captures the inherent geometry of a 3D point cloud and represents it using basic shapes. In Section \ref{3d_classification}, we discuss the usage of ABD-Net as a plug-in network for 3D classification. Implementation details are discussed in Section \ref{experimental_details}. In Section \ref{results}, we present experimental results and analysis of the proposed network. In Section \ref{conclusion}, we provide concluding remarks.

\section{Related works}
\label{section:related_works}
In this section we discuss methods related to our work. We categorize them into 3D point cloud processing methods and 3D point cloud decomposition methods.
\subsection{Point cloud processing methods}
We classify learning based approaches for 3D point cloud processing into view-based, voxel-based and point-based networks.
\subsubsection{View-based networks}
Considering the success of CNNs, one simple way to process 3D point sets is projecting these points onto a 2D image plane to exploit the potential of CNNs \cite{point_trans_nov_9}, \cite{vehicle_view_based}, \cite{autonomous_multiview}, \cite{rotationnet}. The concept here is to transform irregular points to a regular form. In order to consider whole object information, multi-view images are generated which are then passed to 2D-CNNs to extract object features from all directions. This is followed by multi-view feature fusion, where in the network tries to find relations between all views and gives a final 3D object representation. However, because of this transformation, the network only looks at the textural information resulting in loss of the underlying geometric shape information of the 3D point cloud. Thus, these methods are difficult to scale-up for scene understanding or point cloud analysis tasks. Also, the choice of projection plane may heavily influence the recognition performance and occlusion may cause complications.

\subsubsection{Voxel-based networks}
Another approach to transform irregular point set of regular form is 3D voxelization, follwed by 3D convolutions \cite{point_trans_nov_8}, \cite{point_trans_nov_6}, \cite{point_trans_nov_7}. These method establishes enormous amount of memory storage and computational power, as both occupied and non-occupied parts of the scene is represented. There is a cubic growth in the number of voxels which is a function of resolution, making it non-scalable for high resolution data. Some works reduce the complexity of this method by applying convolution only on occupied voxels using sparse convolutions \cite{sparse_convo_1}. However, these methods suffer from loss of local geometry of 3D object due to quantization onto the voxel grid.

\begin{figure*}[ht]
    \centering
    \includegraphics[width=\linewidth]{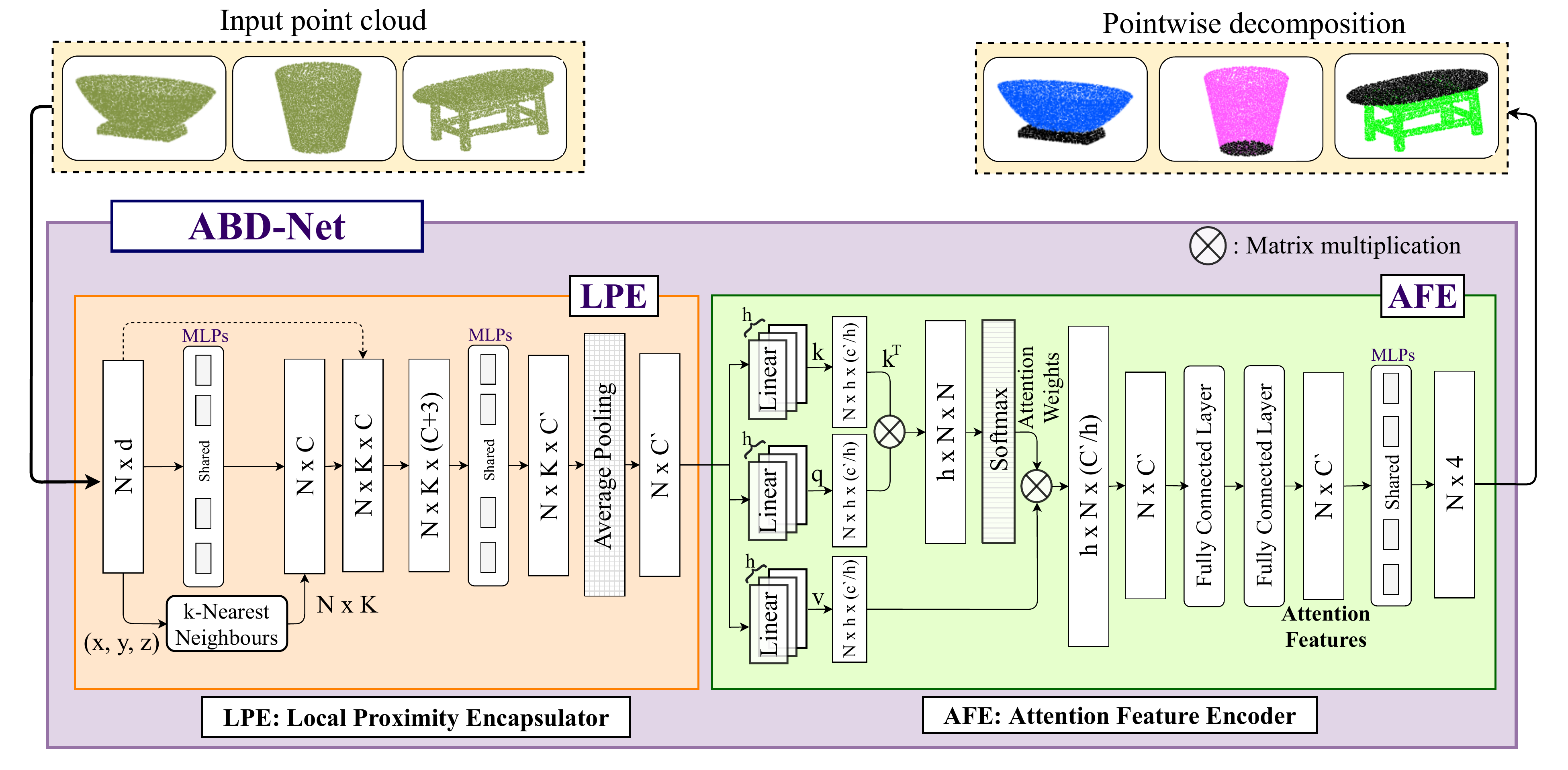}
    \caption{Proposed ABD-Net architecture for 3D point cloud decomposition.}
    \label{fig:ABD-Net}
\end{figure*}

\subsubsection{Point-based networks}
Instead of projecting or quantizing irregular point sets into 2D plane or 3D grids, another method is to directly work of irregular points. PointNet \cite{pointnet} is the first proposed strategy in this approach, where features are learnt for each point followed by a symmetric function, giving a global signature of the point set. However, local context is overlooked in this method. PointNet++ \cite{pointnet++} exploits hierarchical spacial structure, thus learning local geometric layout. The symmetric function mentioned is a universal approximator for any set function \cite{set_pooling_zaheer}. This hinders the capacity of the feature vector to capture important geometric features. We address this issue by modeling relationship between local and global geometric features towards encoding the entire point cloud.

\subsection{Point cloud decomposition methods}
We classify 3D point cloud decomposition methods into two classes, namely classical feature learning and deep feature learning  methods.
\subsubsection{Classical feature learning} Previous works in computer vision for shape decomposition and basic shape detection were performed using RANSAC \cite{ransac} and its variants \cite{ransac_az}, \cite{ransac_jiri}, \cite{ransac_prosac_jiri}. \cite{ransac_efficient}, tries to decompose a point cloud by considering it as a problem of basic shape fitting. \cite{glob_fit} improves on this by optimizing on extracted shapes, based on their relations. However, weakness of RANSAC based approaches are that, the manual parameter tuning is labour-intensive. This demands for careful supervision and makes it non-scable for larger datasets. Another traditional way is to extract hand-crafted features \cite{metric_ten_chris_symbol}, where the authors propose a novel set of hand-crafted features namely metric tensors and christoffel symbols. These features are further used for decomposition using SVM as point classifier. They also show applications of basic shape representation on 3D object super resolution \cite{alatf_super_resolution}, 3D inpainting \cite{altaf_inpaint}, 3D object categorization \cite{metric_ten_chris_symbol}
and 3D object hole filling \cite{altaf_hole_fill}.
We think that performance of this
method may be constrained over the representational power
of the features defined.

\subsubsection{Deep feature learning} Recent advances in deep learning have eliminated the need of hand-crafted features. Many works use deep learning models to extract feature representation of point cloud. \cite{preceptual_shape_decomposition} propose a boundary-based feature extractor, with curvature-based and variation of normal vector constraints, to decompose 3D object into meaningful parts based on perceptual points. Later using this part information, they construct a semantic graph giving explicit shape information. However, this decomposition is highly dependent on the perceptual points selected by the algorithm. \cite{skeleton_decomp} propose to extract curve skeletons with a idea that these can lead to point cloud decomposition. They estimate point normals and local-adaptive thresholds to detect all the possible candidate parts of a point cloud. Then skeleton representations of all optimal parts are predicted followed by linking part skeletons. SPFN \cite{spfn} and ParSeNet \cite{parsenet} propose methods for basic shape fitting to point clouds. \cite{spfn} proposes a supervised method by first predicting per-point segment labels, shape types and normals, and then uses a differentiable module to estimate shape parameters. With this \cite{parsenet} also include B-spline patch as a basic shape and propose a differentiable spline-fitting network.

\section{Attention Based Decomposition Network}
\label{abd-net}

In this section, we discuss the proposed Attention Based Decomposition Network (ABD-Net) for 3D point cloud decomposition as shown in Figure \ref{fig:ABD-Net}. The goal is to represent the inherent geometry of a 3D point cloud using a set of geometric features revealing surfacial information using basic shapes. This is achieved through two modules namely \emph{Local Proximity Encapsulator} (LPE) and \emph{Attention Feature Encoder} (AFE) to extract local and global features using attention based on basic shapes. The first module of our architecture is LPE whose goal is to extract representation for each point using features of its neighborhood.
Next, the AFE learns basic shapes in a point cloud by estimating attention point features which provides global point cloud information.
The learnt attention features are discriminative and can be used for many 3D analysis tasks like 3D classification, hole-filing, upsampling and inpainting. ABD-Net can process input point clouds of various densities.

Consider 3D objects represented as point clouds. Let
$O$ be the set of $M$ point clouds $O = \{P_m\}$, $1 \leq m \leq M$. Let each point cloud $P_m$ contain $N_m$ number of points defined by 3D space point in x, y and z direction, $P_m = \{p_i\}$, $1 \leq i \leq N_m$, where $p_i$ $\in$ $\mathbb{R}^{3}$.

Uniformization theorem \cite{uniformization} says that any 3D object can be decomposed into four basic shapes viz. plane, sphere, cone and cylinder. For our work, we formulate mapping of uniformization theorem as a function of point clouds. To capture the shape information of a 3D point cloud in terms of basic shapes, we propose a function defined as $f: O \rightarrow \Psi$. Here $\Psi$ is set of the same $M$ point clouds with a parameter for basic shape added to each $p_i$ in the set of 3D objects $\Psi = \{P^{'}_m\}$, $1 \leq m \leq M$ and $P^{'}_m = \{p^{'}_i\}$, $1 \leq i \leq N_m$, where $p^{'}_i$ $\in$ $\mathbb{R}^{4}$. The extra dimension in $p^{'}_i$ is the decomposition parameter $l$ indicating the basic shape to which the point $p_i$ belongs to. Here, $l$  takes value 1 for plane, 2 for sphere, 3 for cylinder and 4 for cone. We redefine $\Psi$ as, set of point clouds containing 4 sub sets, $1^{st}$ being planar which contains all the planar points from $M$ point clouds, similarly, $2^{nd}$, $3^{rd}$ and $4^{th}$ for spherical, cylindrical and conical shapes $\Psi = \{\Psi_l\}$, where $1 \leq l \leq 4$. Example, $\Psi_1$ is set of $p_i$ with label 1 corresponding to planar points of all $M$ objects. Similarly for $\Psi_2$, $\Psi_3$ and $\Psi_4$ corresponding to spherical, cylindrical and conical points of all $M$ point clouds respectively.

The input to our model is a point cloud $P_m = \{p_i\}$. The proposed architecture works for $p_i$ $\in$ $\mathbb{R}^{3}$ or $p_i$ $\in$ $\mathbb{R}^{6}$, by considering normals in addition to $X$, $Y$, $Z$ coordinates. For each point cloud the surface information is vital in capturing the geometry which depends on both local as well as global variations. The local and and global geometric variations are captured by LPE and AFE using basic shapes as attention features. The proposed modules LPE and AFE are explained in detail in the following sections.

\subsection{Local Proximity Encapsulator (LPE)}
\label{LPE}

The first module shown in Figure \ref{fig:ABD-Net}, extracts representation of each point considering features of its neighborhood, capturing fine grain details of local point sets. LPE defines patches on point cloud and processes each patch individually extracting local geometric information along with spatial encoding, thus incorporating \emph{local attention}. The module consists of convolution layers (\emph{shared MLPs}) and average pooling layer. The average pooling layer is a symmetric function \cite{pointnet} used to aggregate features along a set of points.

Similar to convolution operations in 2D images that capture spatial variations, these shared weights capture the spatial encoding in 3D point clouds.

LPE initially transforms the points to higher dimensional space using shared MLPs learning spatial encoding of each point $p_i$ by adding $C$ dimensional features. As each point is represented with $C$ dimensional features, we call it ($N \times C$). To account for local geometric information, k-nearest neighbors around each point $p_i$ in Euclidean space $\mathbb{R}^{3}$ are considered giving ($N \times K \times 3$). The number of points required to define the neighborhood vary according to the density of the point cloud. The neighborhood interaction of each point $p_i$ is defined with neighborhood $N_i$ in the local coordinate system of $p_i$ as $p_{ij} = p_{ij} - p_i$.

The neighborhood information from $\mathbb{R}^{3}$ is transferred to higher dimensional feature space for defining neighborhood points in $\mathbb{R}^{C}$, giving ($N \times K \times C$)

To encapsulate these spatial encoding and local geometric information, LPE concatenates local coordinates from $\mathbb{R}^{3}$ and their corresponding feature points in $\mathbb{R}^{C}$ giving ($N \times K \times (C + 3)$).
Each point is further processed using set of shared MLPs, where shared MLPs act as the local feature learners. In order to deal with unordered nature of points in the neighborhood, LPE uses symmetric average pooling function to aggregate the features of the neighborhood.

We set $C = 64$ and $K=32$ and $C`=512$. After each convolution layer we use batch normalization to reduce the covariance shift and use rectified linear unit (ReLU) activation function to add non-linearity to the network for controlling the vanishing gradient problem. After the local geometry with spatial encoding around each point is captured, AFE extracts basic shapes in point cloud and provides global features.

\subsection{Attention Feature Encoder (AFE)}
\label{AFE}

AFE models geometric relationship between the local neighborhood of all the points incorporating \emph{global attention} as shown in Figure \ref{fig:ABD-Net}. This module takes in the local neighborhood information provided by LPE, and extracts global features of point cloud $P_m$ which are attention features. 
The attention mechanism used here is known as "Scaled Dot-Product Attention" given by:
\cite{attention_is_all_you_need}
\begin{equation} \label{att_equ}
    Attention(Q, K, V) = softmax(\frac{QK^T}{\sqrt{d_k}})V
\end{equation}
where Q, K and V are \emph{queries}, \emph{keys} and \emph{values} matrices and $d_k$ is dimension of \emph{keys}. Here, Q, K and V are abstractions of the input in transformed space. The query matrix represents the target input point which is to be processed, key matrix represents key features of the input and value matrix is a representation of the input. Attention mechanism has been first proposed and widely used in NLP tasks \cite{attention_is_all_you_need} to give importance to selected words depending on the language context. Here, we aim to bring attention mechanism in point clouds to model relationship between the neighborhoods of all the points resulting in capturing global point cloud information. The attention features are extracted based on basic shapes in a point cloud. The idea is whenever we are required to calculate the attention of a target point with respect to the input points, we should use the query of the target and the key of the input to calculate a \emph{score}. 
The score is then multiplied with the value matrix to keep intact the values of the points we want to focus on, and diminish irrelevant points. The attention mechanism is performed in different representational sub-spaces, each sub-space is referred as head. AFE consists of two sub-layers. The first is a multi-headed attention layer and the second is a fully connected feed-forward network. 
Attention mechanism is used to find the set of points that should influence the target encoding of the query point. AFE is equipped with \emph{multiple} attention mechanism (termed as \emph{head}: $h$) to directly model geometric relationships between all the points in a point cloud in different \emph{representational sub-spaces}, regardless of their respective position. 

AFE first embeds the local features provided by LPE in 3 different spaces to get respective query, key and value matrices using $3$ independent linear layers. These linear layers learn the transformation from the local feature space to Q, K, V spaces. The dot-product of query and key matrices are passed to softmax function to generate attention weights which are further multiplied with value matirx to get attention features. The following process is performed parallelly across multiple heads dealing with different representational sub-spaces. The multi-headed attention features are then refined by a set of fully connected layers to output global attention features.
The attention features captures the inherent geometry of a 3D point cloud and represents it using basic shapes namely, plane, sphere, cone and cylinder which is vital for 3D shape analysis. Therefore attention features can be used as geometrical features for the various 3D analysis task like 3D classification, 3D hole-filling and 3D upsampling.

We use $3$ Attention Feature Encoders connected consecutively in our model architecture. In each AFE, we set $h=4$. After each fully connected layer in our network, we use batch normalization and Rectified Linear Unit (ReLU) activation function. LPE and AFE together capture local and global geometric variations using basic shapes as attention features. The attention features can be used for classification of 3D objects.

\begin{figure}
    \centering
    \includegraphics[width=\linewidth]{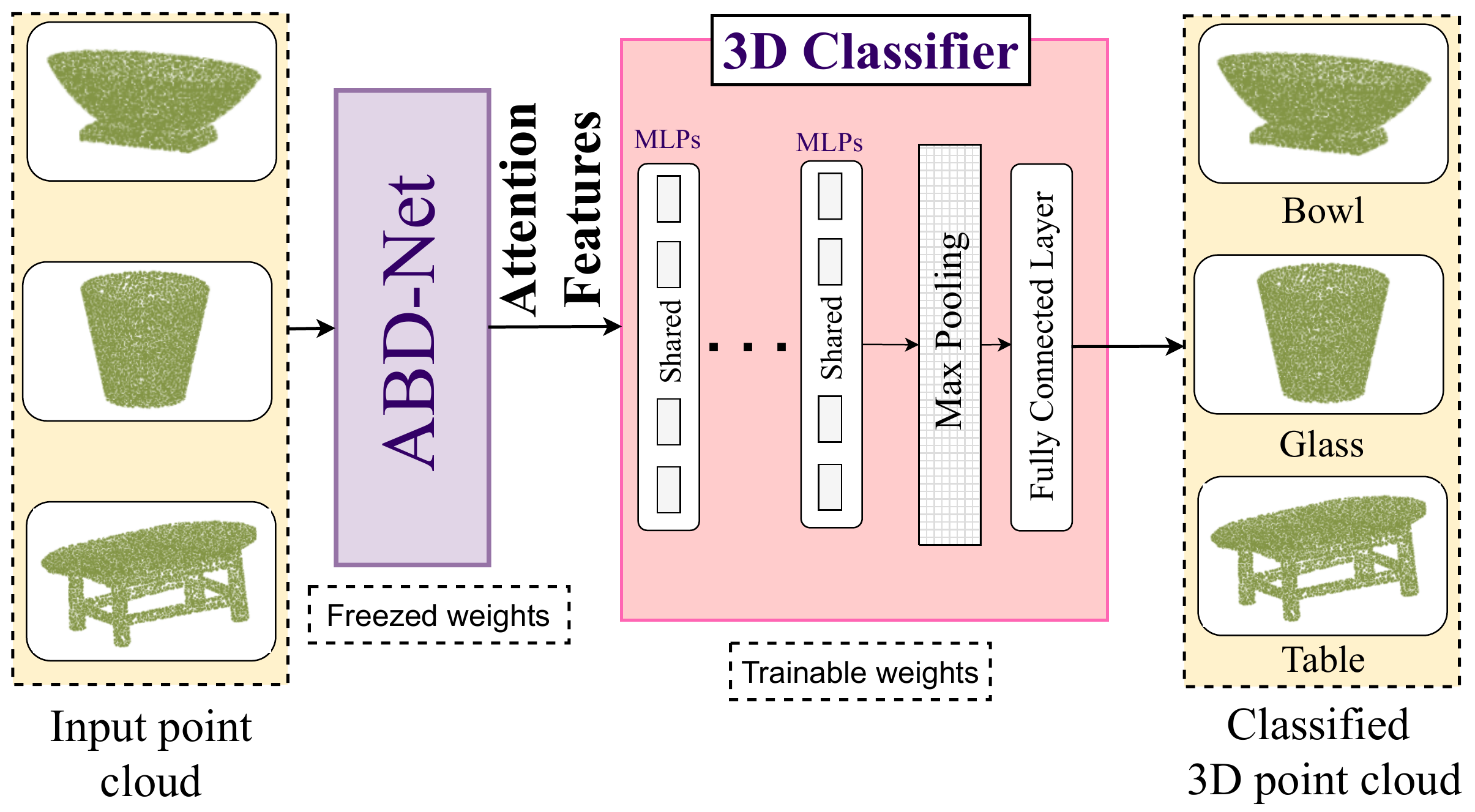}
    \caption{ABD-Net + 3D Classifier.}
    \label{fig:ABD-net + Simple 3D Classifier}
\end{figure}

\begin{figure*}
    \centering
    \includegraphics[width=\linewidth]{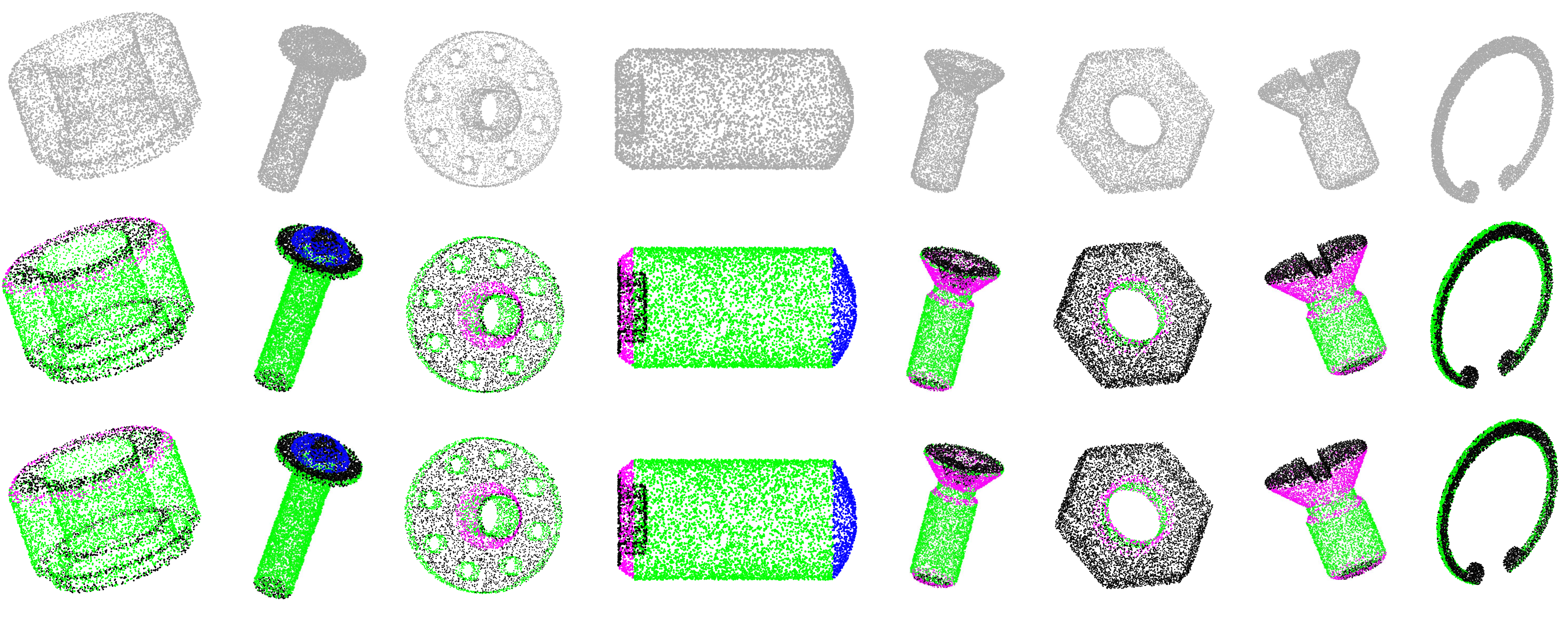}
    \caption{Visualization of results of basic shape decomposition using our proposed ABD-Net architecture on ANSI mechanical components dataset. The top row shows the original point clouds, the middle row shows the ground-truth point cloud decomposition, and the bottom row shows the decomposition result of our architecture. The  points in black represents planar shape, the points in blue represents spherical shape, the points in green represents cylindrical shape and points in magenta represents conical shape.}
    \label{fig:trace}
\end{figure*}

\begin{figure*}
    \centering
    \includegraphics[width=\linewidth]{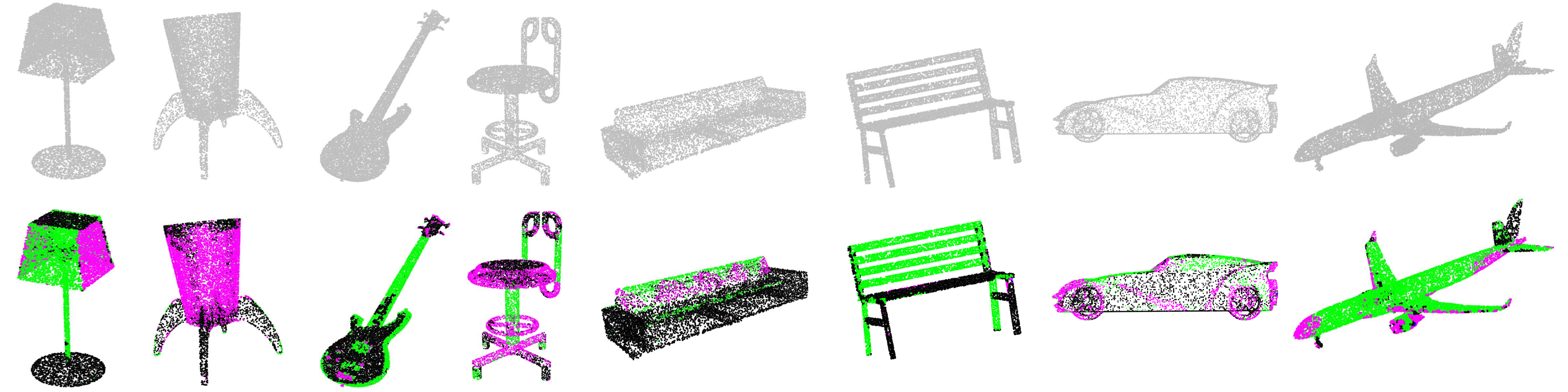}
    \caption{Visualization of results of basic shape decomposition using our pre-trained ABD-Net on ModelNet40 dataset. The top row shows the original point clouds and the bottom row shows the decomposition result of our architecture. The color coding of decomposition remains the same as that of results shown in the Figure \ref{fig:trace}. Even though ABD-Net is trained on ANSI dataset, these results show generalizability for decomposition on completely diverse set of objects from ModelNet40.} 
    \label{fig:model}
\end{figure*}

\begin{figure*}
    \centering
    \includegraphics[width=\linewidth]{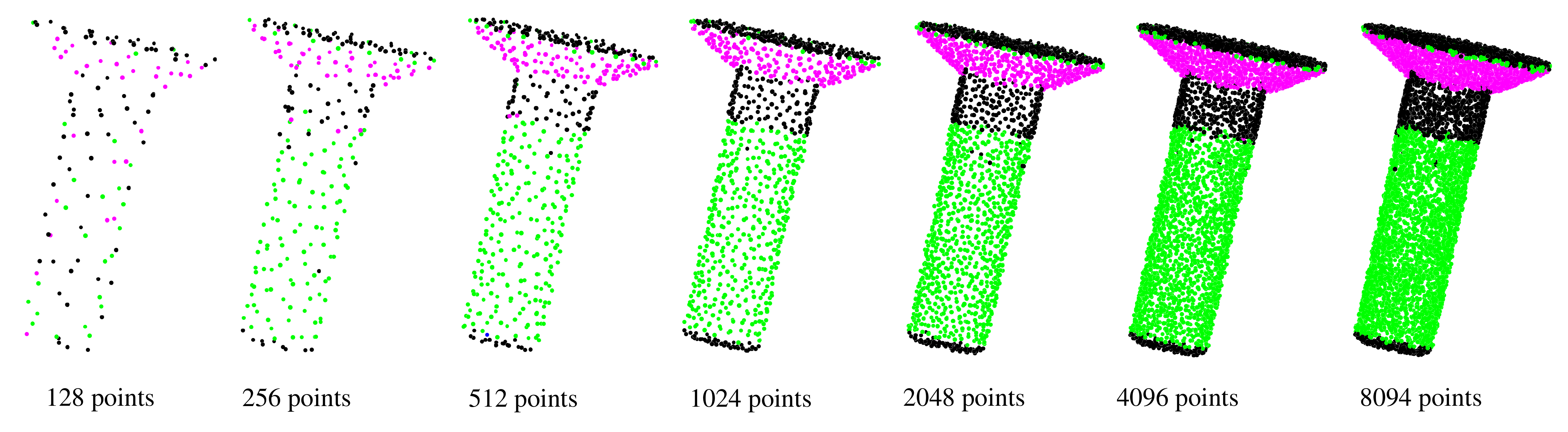}
    \caption{Visualization of results of basic shape decomposition using proposed ABD-Net on a sample point cloud from ANSI dataset with varying point density. The color coding of decomposition remains the same as that of results shown in the Figure \ref{fig:trace}.}
    \label{fig:single_density}
\end{figure*}

\section{3D classification}
\label{3d_classification}
The ABD-Net is trained only on ANSI mechanical components dataset which has 3D point clouds with basic shape label assigned to each point. ABD-Net learns attention features based on basic shapes present in 3D point clouds of ANSI mechanical components dataset. Representing a point cloud with basic shapes can improve the performance of 3D classification. The classification of 3D objects is carried out on ModelNet40 dataset with the attention features acquired by pre-trained ABD-Net. Similar to transfer learning, the weights of ABD-Net are freezed while training the 3D classifier. A set of 4 shared MLPs are used to classify 3D objects based on their attention features. To each MLP in the 3D classifier we additionally provide the point coordinate information of the input point cloud.

\section{Experimental details}
\label{experimental_details}
In this section we discuss about the dataset used for training our ABD-Net and 3D object classifier, with their implementation details while training. 

\subsection{Datasets}
We use American National Standards Institute
mechanical component dataset, provided by Traceparts \cite{traceparts} to train our ABD-Net. It includes 3D models of mechanical tools such as nuts, bolts with basic shape labels, as shown in Figure \ref{fig:trace}. We use a train/test split of $12984$/$3172$ respectively. The categories are different in both sets, making training and testing sets disjoint. Each object has $8096$ points, with their coordinates and normals. The associated groud truth basic shape labels for each object in the dataset is provided by Traceparts. As data preparation procedure, we uniformly sample $1024$ points from each point cloud with the associated normal vectors. We keep normal vectors as optional additional feature for training ABD-Net. We exclude the point clouds from train and test set, having more than 90\% planar shape category to prevent dataset skewness towards planar shape.

We use ModelNet40 \cite{point_trans_nov_7} dataset to train a 3D classifier with pre-trained ABD-Net as a pre-processing unit. ModelNet40 consists of $12,311$ CAD models with a total of $40$ categories, where $9,843$ objects are used for training and $2,468$ for testing. As data preparation procedure, we uniformly sample $1024$ points from each CAD model with the normal vectors from the object meshes. We keep normal vectors as optional additional feature for training 3D object classification.

\subsection{Implementation details}
During training both decomposer and classifier, we augment the networks input by random rotation, scaling. In addition to these augmentation we also use random points dropout for classifier training. We train our decomposer network for $50$ epochs and classifier network for $200$ epochs. For training, we use Adam optimizer \cite{adam}  with batch size $16$ and learning rate $0.001$ with learning rate decay of $0.5$. We train our decomposer ABD-Net on NVIDIA Corporation GV100GL [Quadro GV100] 230 Volta GPU with 32GiB memory and classifier on NVIDIA GeForce RTX 3090 ICHILL X4 GPU with 24GiB memory. Both the networks are implemented in PyTorch framework \cite{pytorch}.

\section{Results and discussions}
\label{results}
In this section, we show the results of proposed ABD-Net architecture using ANSI mechanical components and ModelNet40 dataset. We also compare the results of 3D object classification with state-of-the-art techniques and show comparable results.

\subsection{Shape decomposition}
We show decomposition results on ANSI mechanical components dataset in Figure \ref{fig:trace} and on ModelNet40 dataset in Figure \ref{fig:model}. Our proposed ABD-Net achieves an overall accuracy of $99.3\%$ for basic shape decomposition on ANSI test set. In Figure \ref{fig:trace}, we can see that there is clear demarcation at the edges of all objects, demonstrating the ability of our model to predict inherent shape of the model by looking at its surface. For objects from ModelNet40 as shown in Figure \ref{fig:model}, the transition between basic shapes is smooth, thus increasing the difficulty of decomposition. We can observe that, our model retains its decomposition performance even when the surface complexity of the objects increases. Also, the object shapes are totally diverse as compared to the object on which our ABD-Net is trained on, showing generalizability for decomposition. We provide ablation study for shape decomposition in Section (\ref{ablation}).

\begin{center}
\begin{table}
\caption{Results of 3D object classification of ModelNet40 benchmark dataset and comparison with state-of-the-art techniques with $1024$ point cloud density (nor: normal).}
\begin{tabular}{ l| c| c| c} 
\hline
Method & input  & \#params & acc.\\
\hline
 PointNet & xyz & 3.50M & 89.2\\
 PointNet++ & xyz  & 1.48M & 90.7 \\
 KCNet & xyz  & - & 91.0 \\
 MRTNet & xyz  & - & 91.2 \\
 Spec-GCN & xyz  & - & 91.5 \\
 Spec-GCN  & xyz, nor  & - & 91.8 \\
 \textbf{$\ast$} 3D Classifier & xyz, nor  & 500K & 92.1 \\
 DGCNN & xyz  & - & 92.2 \\
 PCNN & xyz  & 8.20M & 92.3\\
 PointWeb & xyz  & - & 92.3\\
  \textbf{$\ast$ ABD-Net+3D classifier}& \textbf{xyz}  & \underline{\textbf{500K}} & \underline{\textbf{92.2}} \\
 PointConv & xyz, nor  & - & 92.5 \\
 Point Transformer & xyz, nor  & 13.5M & 92.8 \\
 \textbf{$\ast$ ABD-Net+3D classifier}& \textbf{xyz, nor}  & \underline{\textbf{500K}} & \underline{\textbf{92.8}} \\
 RSCNN & xyz  & 1.41M & 92.9 \\
 PCT & xyz  & 2.88M & 93.2 \\
 \textbf{PointTransformer} & \textbf{xyz}  & - & \textbf{93.7} \\
 \hline
\end{tabular}
\label{table:table1}
\end{table}
\end{center}

\subsection{Shape classification} 
We use a 3D classifier with 4 shared MLPs followed by a max-pooling layer, with a pre-trained ABD-Net for point cloud decomposition as a pre-processing step. The trainable weights of pre-trained ABD-Net are freezed while training this classifier.
We compare the performance of the classifier with and without the decomposition method, demonstrating the effectiveness of ABD-Net. Table \ref{table:table1} shows improved classification performance using our pre-prained ABD-Net as a plug-in network before 3D classifier. We can observe, that the classification accuracy of 3D classifier increase from $92.1$\% to $92.8$\% by incorporating ABD-Net as a pre-processor. This shows that, the shape decomposition features are well exploited by the 3D classifier increasing the classification performance. This also implies, that the basic shape representation of a point cloud is well suited for better 3D visual analysis tasks.

The quantitative comparisons with the state-of-the-art techniques is shown in Table \ref{table:table1}. Our proposed classifier achieves improved results over many techniques. Point transformer \cite{zhaodec2020point} based classifiers are current state-of-the-art techniques that directly work on raw point cloud. However, the number of trainable parameters in these methods are quite high. The 3D classifier achieves $92.8\%$ overall classification accuracy on ModelNet40 dataset with 500K trainable parameters, which is $3$ times less than the other methods.

\subsection{Ablation study}
\label{ablation}
\subsubsection{Robustness test for shape decomposition}
To test the robustness of our proposed ABD-Net for point cloud decomposition, we perform point density variation and point perturbation test as a part of ablation study.
\\
\begin{figure}
    \centering
    \includegraphics[width=\linewidth]{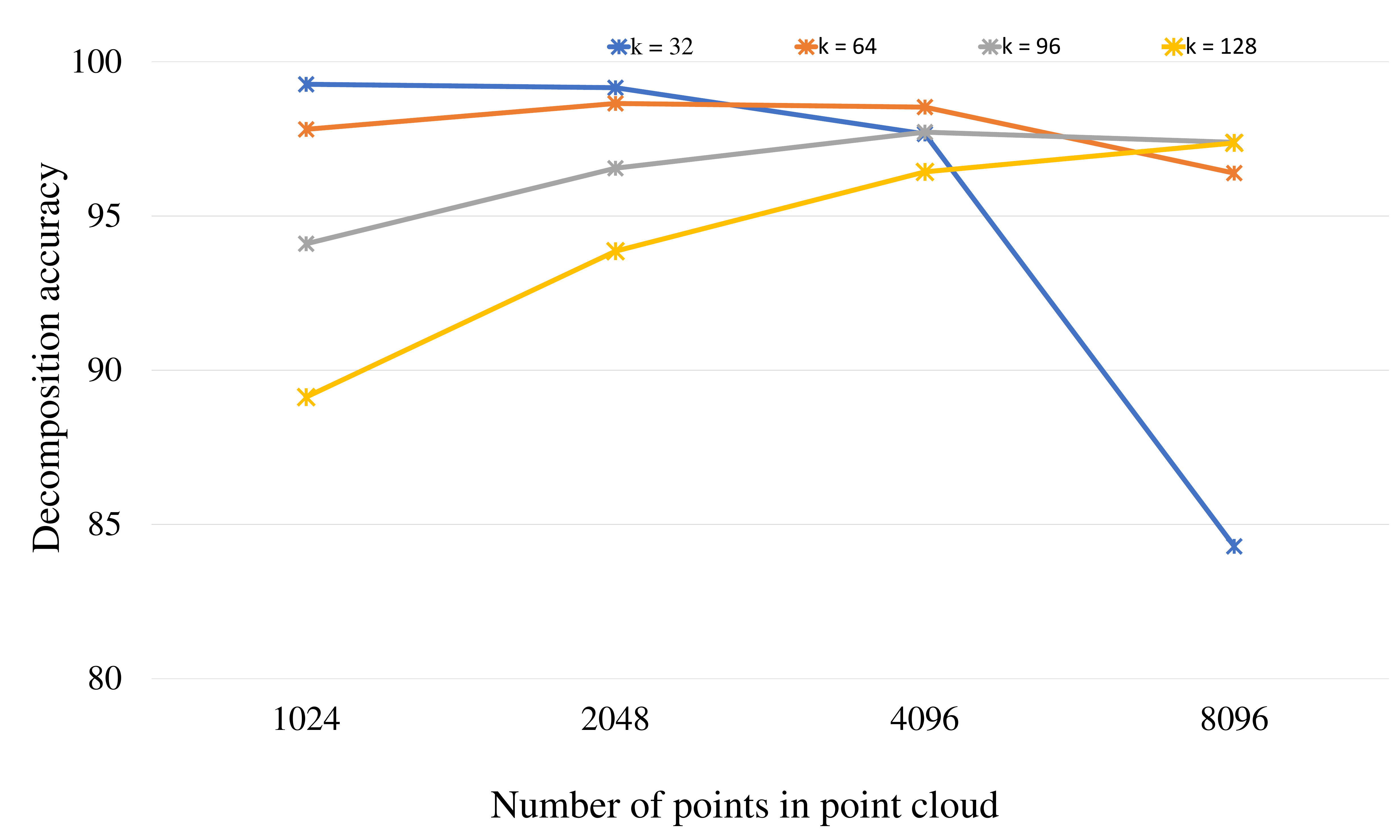}
    \caption{Decomposition results of ABD-Net for varying point density and neighborhood points (k) on ANSI Dataset. The graph depicts the dependency of number of neighbourhood points ($k$) on density of the input point cloud.}
    \label{fig:density full data}
\end{figure}

\begin{figure*}
    \centering
    \includegraphics[width=\linewidth]{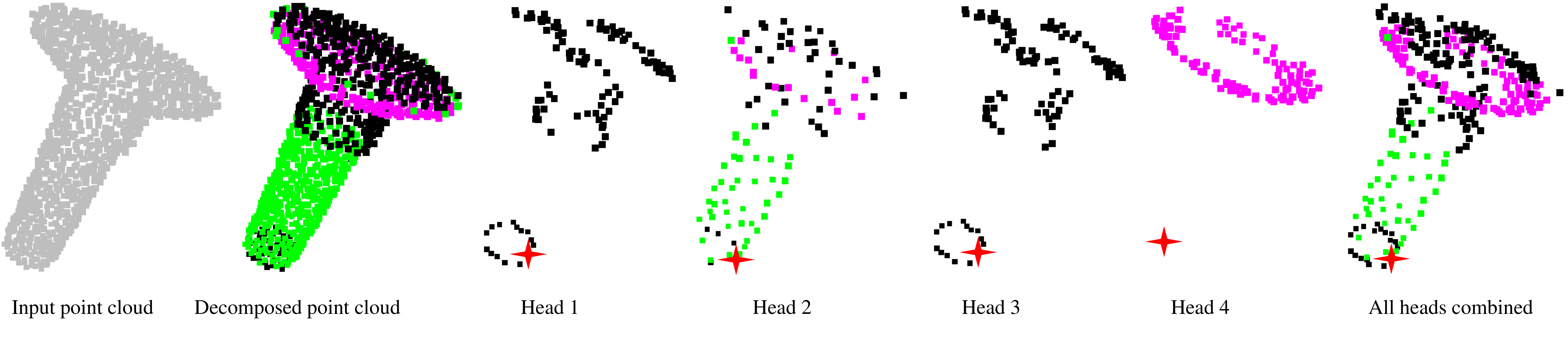}
    \caption{Visualization of the attention weights extracted by four heads, from the third AFE module. The star represents the query point for which attention is computed.}
    \label{fig:attention_visualization}
\end{figure*}

\hspace{-0.35 cm}\textbf{Affect of neighborhood.} In Figure \ref{fig:density full data}, we show an analysis of point density and neighborhood points (k) variation against decomposition accuracy on whole ANSI test dataset. We perform this experiment with a pre-trained ABD-Net on 1024 points with $k=32$. We show the dynamic nature of ABD-Net over number of points that it can process as input. We can observe that, with $k=32$ the decomposition accuracy starts dropping as the point density increases. This behaviour is observed because, as point density increases the spread of neighboring points over the surface starts decreasing, making the defined local patches too small to capture local geometrical information. To handle this, a simple way is to increase the number of points defining a neighborhood as the density of points in a point cloud increases. Also, decomposition accuracy is low when $k=128$ with 1024 number of points in a point cloud. This is observed, as the patch defined exceeds the local context, and ABD-Net is wrongly influenced by global features. In Figure \ref{fig:density full data}, we can observe this intuitive pattern of accuracy against point density and neighborhood points as explained. We use $k$ as $32$, $64$, $96$ and $128$ for point clouds with $1024$, $2048$, $4096$ and $8096$ points.\\

\begin{figure}
    \centering
    \includegraphics[width=\linewidth]{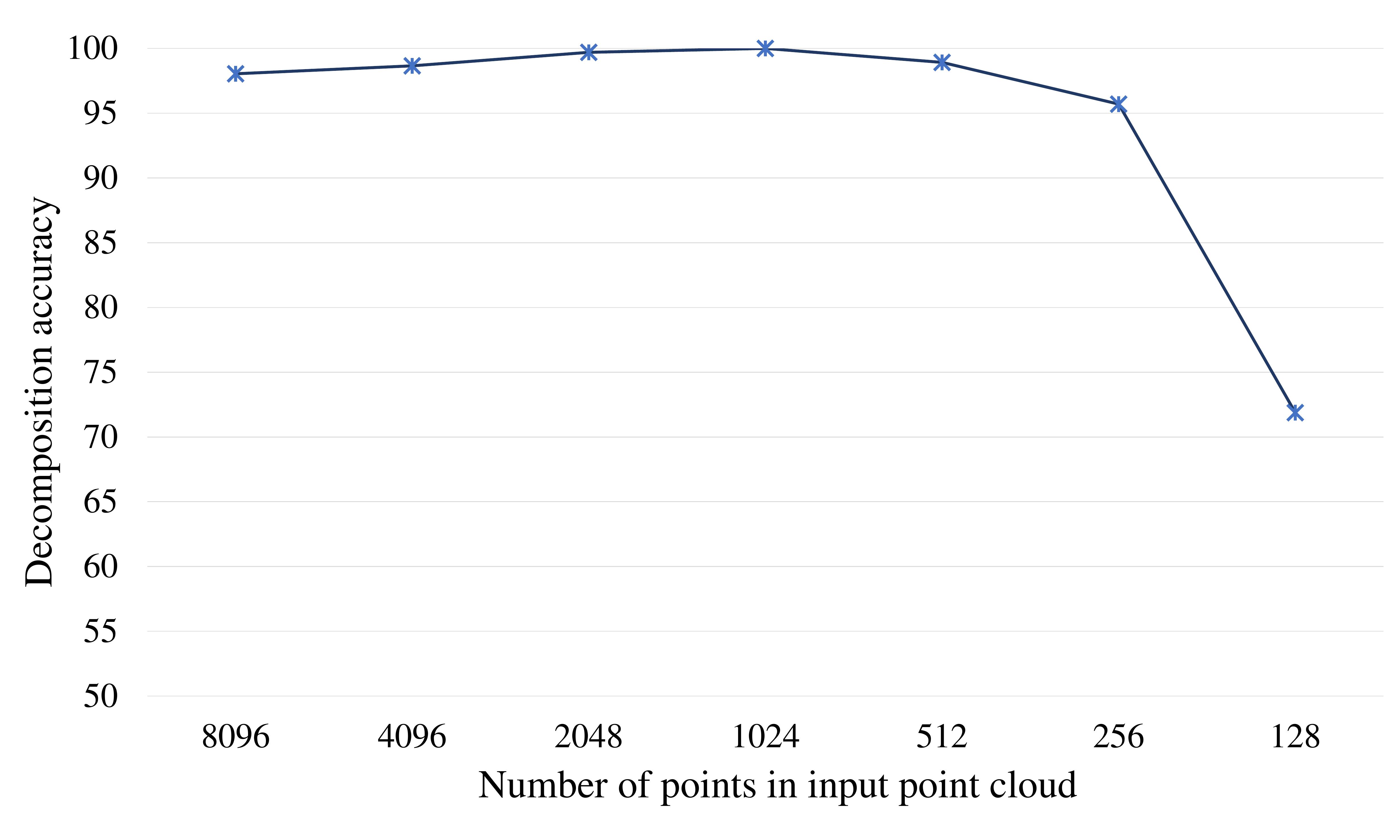}
    \caption{Decomposition analysis of ABD-Net for varying point density on sample point cloud from ANSI dataset shown in Figure \ref{fig:single_density}. This graph depicts that even though ABD-Net is trained on $1024$ points, it maintains good performance on varying densities of input point cloud.}
    \label{fig:graph_single_density}
\end{figure}

\hspace{-0.35cm} \textbf{Affect of density.} We sample $128$, $256$, $512$, $1024$, $2048$, $4096$ and $8096$ as shown in Figure \ref{fig:single_density}, and demonstrate robust decomposition by ABD-Net for varying sampling density. Figure \ref{fig:graph_single_density}, shows the instance decomposition accuracy of our proposed ABD-Net on a sample from ANSI dataset shown in Figure \ref{fig:single_density}. We can observe that decomposition accuracy is $100\%$ when number of points is $1024$. It also maintains good performance for $8096$, $4096$, $2048$, $512$ and $256$ point densities. However, we observe a $23.8\%$ drop in decomposition accuracy with point density as $128$. This is observed because, with increase in sparsity of the point clouds, there is proportional increase in difficulty for surface prediction and thus, increasing the difficulty for basic shape decomposition. Even though our ABD-Net is trained on $1024$ points, it manages to keep up its performance with varying densities exhibiting its density-invariant property.\\

\begin{figure}
    \centering
    \includegraphics[width=\linewidth]{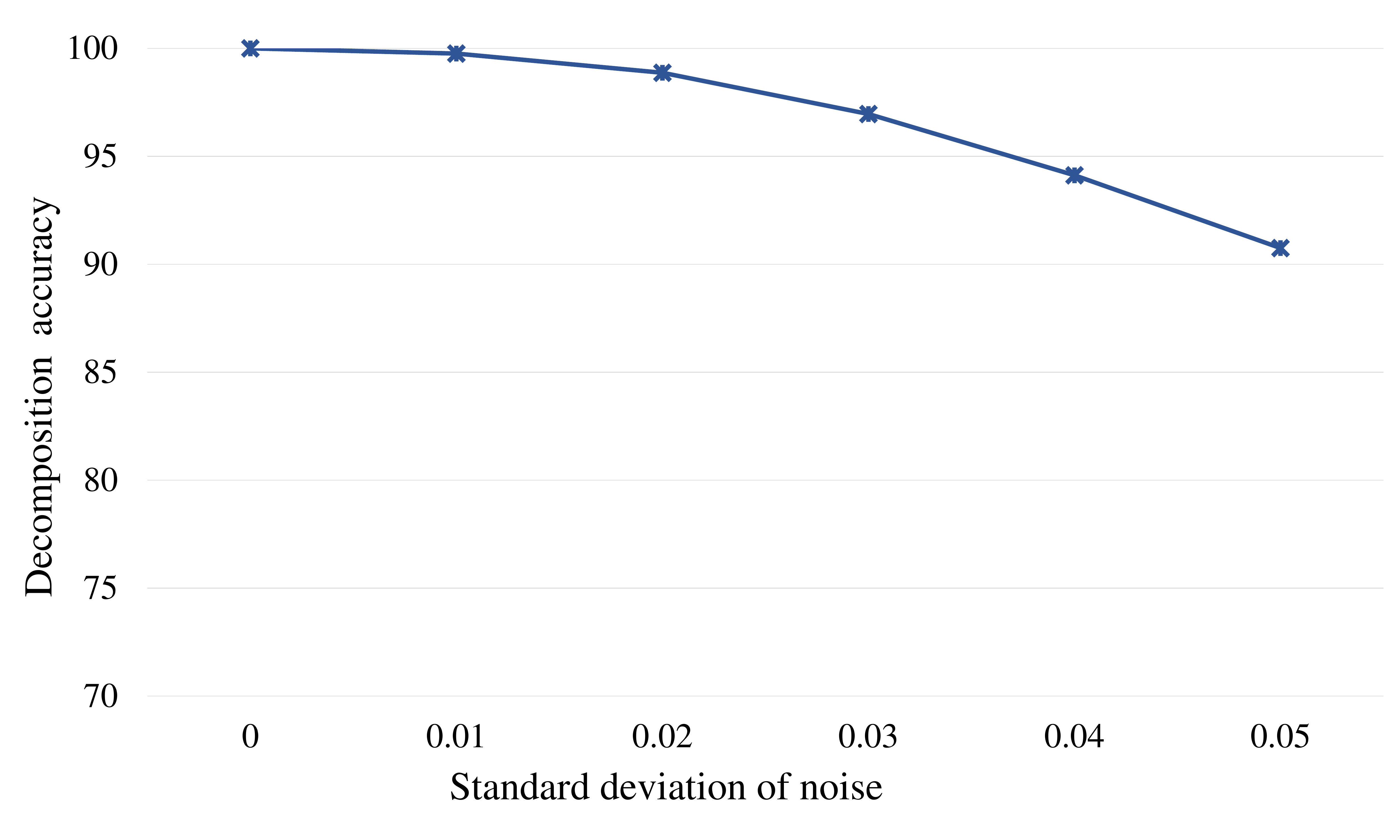}
    \caption{Decomposition analysis of ABD-Net for various degrees of point perturbation on sample point cloud from ANSI dataset shown in Figure \ref{fig:single_noise}. The graph depicts good performance of ABD-Net even with rigorous point perturbations.}
    \label{fig:graph_single_noise}
\end{figure}

\begin{figure}
    \centering
    \includegraphics[width=\linewidth]{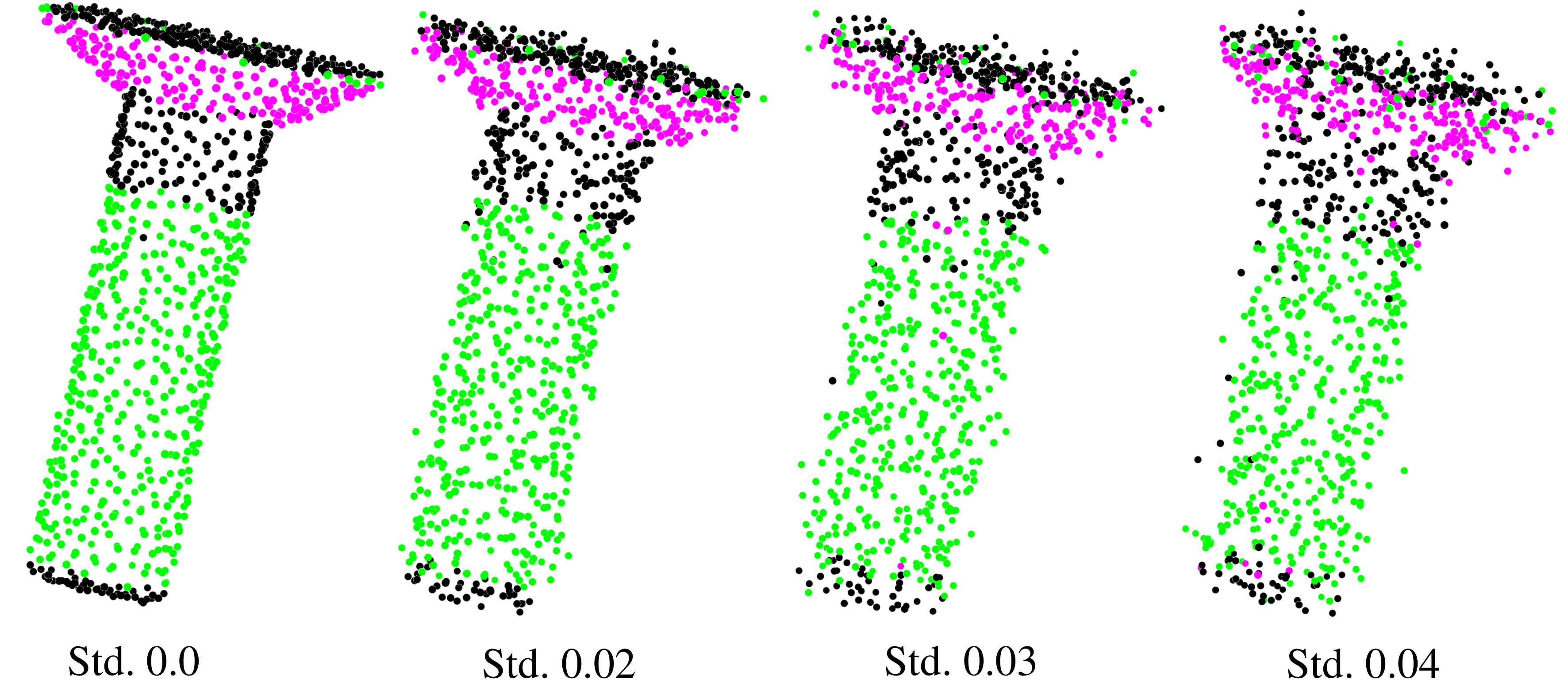}
    \caption{Visualization of results of basic shape decomposition using proposed ABD-Net on a sample point cloud from ANSI dataset with various degrees of point perturbations.}
    \label{fig:single_noise}
\end{figure}

\hspace{-0.35 cm}\textbf{Point perturbation.} 
We also show decomposition performance of our proposed ABD-Net with input point perturbations. Gaussian noise is randomly added to each point independently as shown in Figure \ref{fig:single_noise}, with standard deviation of noise being $0.0$, $0.02$, $0.03$, $0.04$ respectively. Figure \ref{fig:graph_single_noise}, shows quantitative analysis of decomposition accuracy with addition of noise to input point cloud having 1024 points. Our ABD-Net achieves an decomposition accuracy of about $91\%$ even when the point clouds are distorted with severe noise with a standard deviation of $0.05$. Similar to point cloud sparsity, with increase in point cloud distortion, there is proportional increase in surface prediction making the task of basic shape decomposition difficult. These results indicate that our proposed ABD-Net is robust to point distortions, thus exhibiting its noise-invariant property.

\begin{figure}
    \centering
    \includegraphics[width=\linewidth]{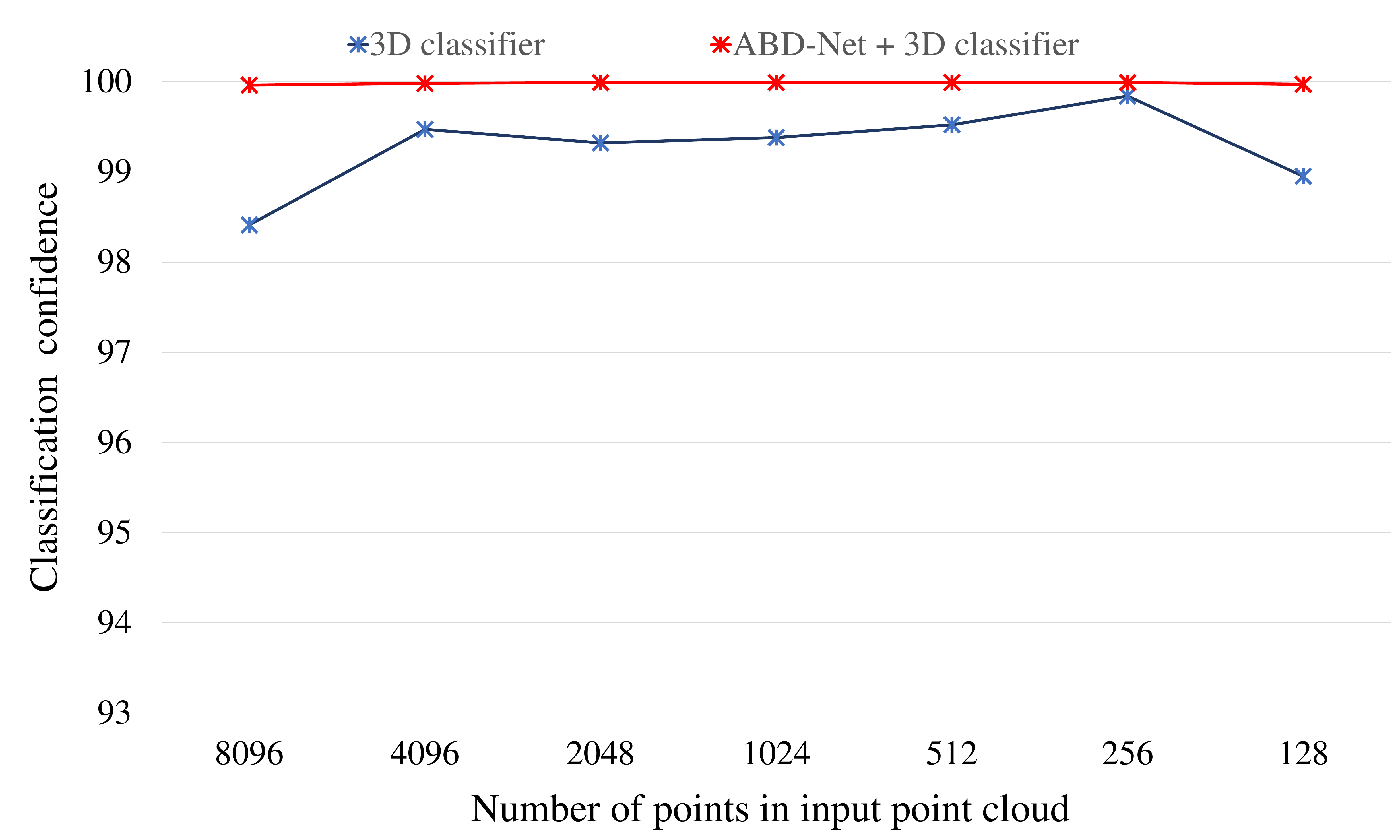}
    \caption{Comparison of density variation test between 3D classifier and ABD-Net + 3D classifier on random sample point cloud from airplane class in ModelNet40. This graph shows constant better performance in classification by ABD-Net + 3D classifier for different density input point clouds.}
    \label{fig:classification_density_comparision}
\end{figure}

\begin{figure}
    \centering
    \includegraphics[width=\linewidth]{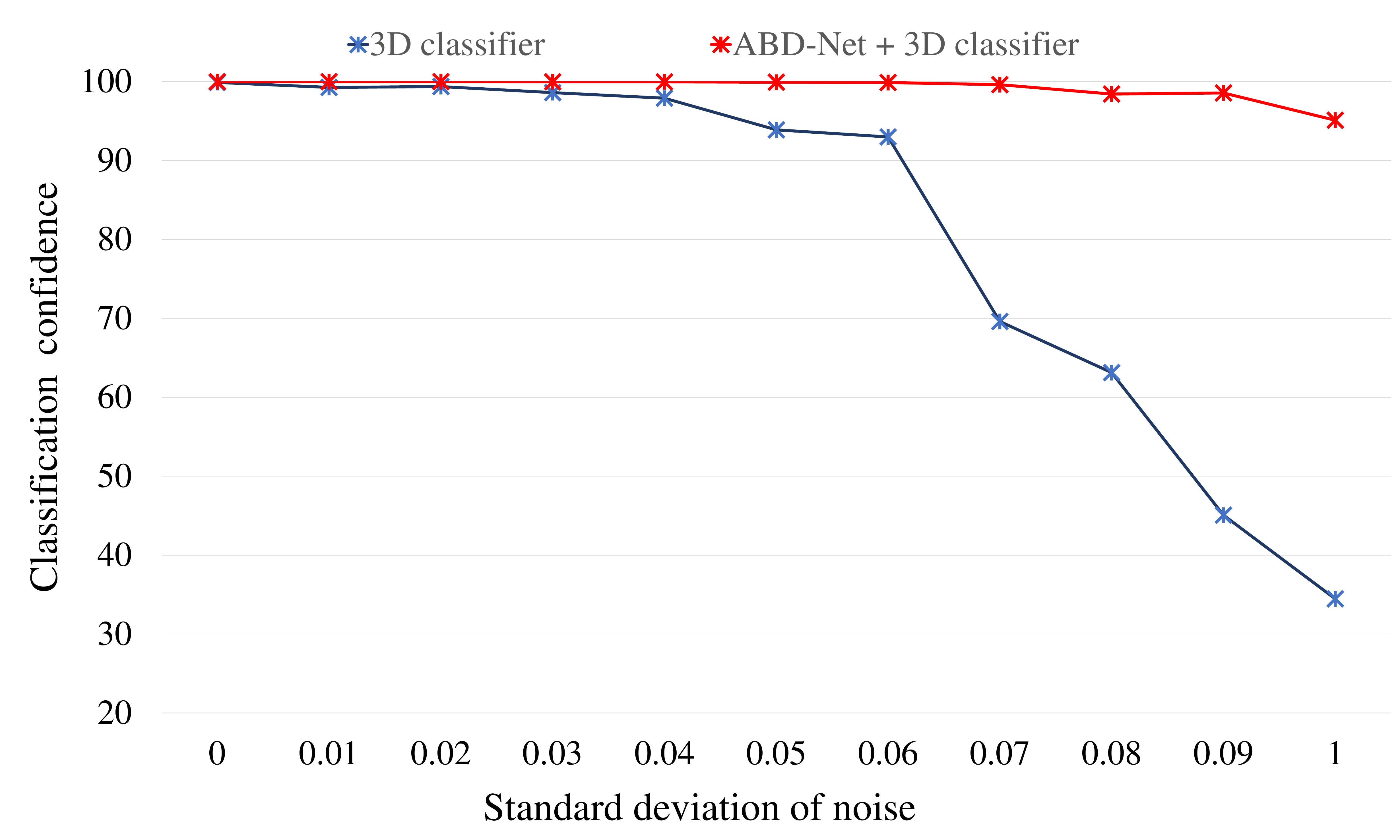}
    \caption{Comparison of point perturbation test between 3D classifier and ABD-Net + 3D classifier. This graph depicts remarkable increase in classification performance using ABD-Net as a pre-processing step for 3D classifier with rigorous point perturbation.}
    \label{fig:classification_comparision_with_noise_variation}
\end{figure}

\subsubsection{Attention visualization}
In Figure \ref{fig:attention_visualization}, we visualize the attention weights given by our attention mechanism in AFE for a target (query) point. We call these attention as the importance score given by the our model to all the points. The query point embedding is computed by considering each point, in accordance with its importance. We show the top $100$ points selected for the query point, extracted by each head from the third AFE module. Figure \ref{fig:attention_visualization} shows a sample 3D point cloud from ANSI dataset and its corresponding basic shape decomposition with $100$\% instance accuracy, followed by the visualization of attention points from each head. We can observe that each head models different kind of importance relationship. The \emph{head} $1$ is trying to capture points having planar property, where as \emph{head} $2$ is trying to capture an overall geometry by relating points from each basic shape. The  \emph{head} $3$ refines the planar points captured by \emph{head} $2$. At last \emph{head} $4$ captures points having conical property. In this way all the head combinedly capture the whole point cloud shape information, thus incorporating \emph{global attention}.

\subsubsection{Robustness test for shape classification}
We demonstrate the robust performance of our proposed architecture ABD-Net by showing the classification accuracy on variation in input point clouds.\\

\hspace{-0.35cm}\textbf{Point density variation.} 
In Figure \ref{fig:classification_density_comparision}, we show an analysis of point density variation. It shows instance classification confidence of 3D classifier and 3D classifier with a pre-trained ABD-Net as pre-processor (ABD-Net + 3D classifier). This experiment is done on a random sample 3D point cloud from airplane class in ModelNet40 dataset. We sample $8096$, $4096$, $2048$, $1024$, $512$, $256$ and $128$ points and show performance of both the classifiers. We can observe that, the classification confidence remains constant of ABD-Net + 3D classifier, where as 3D classifier alone struggles with varying point densities. This implies that the basic shape representation of the point cloud acquired by ABD-Net is better for classification task. Also, the extracted significant features are not affected by point point density, which helps for better classification.\\

\hspace{-0.35cm}\textbf{Point perturbation.}
In Figure \ref{fig:classification_comparision_with_noise_variation}, we show an analysis of point perturbation. It shows instance classification confidence of 3D classifier and 3D classifier with a pre-trained ABD-Net as pre-processor (ABD-Net + 3D classifier). We use the same point cloud which was used for the density variation test. Gaussian noise is randomly added to each point independently, with standard deviation of noise varying from 0.01 to 1.0. We can observe that even with severe point cloud distortion with a noise having standard deviation of 1.0, ABD-Net + 3D classifier performs exceedingly well than that of 3D classifier alone. The 3D classifier starts to struggle when standard deviation of noise approaches to 0.06, where as ABD-Net + 3D classifier maintains its confidence score above 94\% at all noise levels. This implies that the basic shape representation of the point cloud acquired by ABD-Net is in-variant to noise which improves the classifier performance to a large extent.

\section{Conclusion}
\label{conclusion}
In this paper, we have proposed ABD-Net, a deep architecture that captures the inherent geometry of a 3D point cloud and represents it using basic shapes namely, plane, sphere, cone and cylinder. The proposed model contains LPE to capture local geometry with spatial encoding around each point. The next module in ABD-Net is AFE to learn basic shapes in point cloud using attention features based on basic shapes. AFE models geometric relationship between the neighborhoods of all the points resulting in capturing global point cloud information. We demonstrated the results of the proposed ABD-Net on ANSI mechanical components dataset and ModelNet40 dataset. Further, we have also shown that the basic shape representation acquired by ABD-Net is better for 3D classification task. We have demonstrated improved classification results of using attention features acquired by proposed ABD-Net and compared with other classification methods.

\bibliographystyle{IEEEtran}
\bibliography{mybib}

\end{document}